\documentclass[11pt]{article}

\usepackage[preprint]{acl}

\usepackage{times}
\usepackage{latexsym}
\usepackage{amsmath}
\usepackage{multicol}
\usepackage{hyperref}

\usepackage[T1]{fontenc}

\usepackage[utf8]{inputenc}

\usepackage{microtype}

\usepackage{inconsolata}

\usepackage{graphicx}

\title{ReLeVAnT: Relevance Lexical Vectors for Accurate Legal Text Classification}

\author{
\begin{tabular}{c @{\hspace{4em}} c}
Ishaan Gakhar & Harsh Nandwani \\[0.0em]
Perssonify & Perssonify \\[0.0em]
 Levi \& Korsinsky & Levi \& Korsinsky \\
\texttt{ishaan@perssonify.com} & \texttt{harsh@perssonify.com} \\
\\[0.6em]
\multicolumn{2}{c}{\texttt{research@perssonify.com}}
\end{tabular}
}

\begin{document}
\maketitle
\begin{abstract}

The classification of legal documents from an unstructured data corpus has several crucial applications in downstream tasks. Documents relevant to court filings are key in use cases such as drafting motions, memos, and outlines, as well as in tasks like docket summarisation, retrieval systems, and training data curation. Current methods classify based on provided metadata, LLM-extracted metadata, or multimodal methods. These methods depend on structured data, metadata, and extensive computational power. This task is approached from a perspective of leveraging discriminative features in the documents between classes. The authors propose ReLeVAnT, a framework for legal document binary classification. ReLeVAnT utilises n-gram processing, contrastive score matching, and a shallow neural network as the primary drivers for discriminative classification. It leverages one-time keyword extraction per corpus, followed by a shallow classifier to swiftly and reliably classify documents with 99.3\% accuracy and 98.7\% F1 score on the LexGLUE dataset.

\end{abstract}

\section{Introduction}

Binary classification of legal documents based on relevance is a multi-faceted task. Keeping downstream tasks in mind, the authors approach this problem statement from a robustness, coverage and computationally inexpensive perspective. Here, defining `relevance' is key; within the scope of this work, it is considered to be linked to court filings and relatedness to proceedings. These relevant documents can be leveraged in the legal domain for a vast number of purposes, including legal workflow automation \cite{grossman2010technology}, docket summarization \cite{bhattacharya2021incorporating} \cite{saravanan2008automatic}, retrieval systems \cite{chalkidis2022lexglue} \cite{pipitone2024legalbench}, litigation surveillance \cite{ashley2017artificial} \cite{katz2017general} and training data curation for larger ML systems \cite{chalkidis2019large} \cite{zhong2020does}.

Most prevalent methods have limitations with respect to the type of documents, computational cost, or assumptions of access to metadata. Where works like \citet{undavia2018comparative} achieve 72.4\% accuracy in the classification of SCOTUS cases into legal issues, their framework is topic-based and not fact-based, resulting in text being required in opinion formats. Similarly, \citet{li2025document} achieves impressive results on the mentioned dataset with remarkable speed; the proposed method requires indicative filenames and fails to handle out-of-scope documents, proving unsuitable for noisy naming conventions, often encountered when dealing with massive corpora.  A substantial improvement of upto 20.6\% in the accuracy of long legal document classification by \citet{limsopatham2021effectively}, however, the computational cost is very high and makes it infeasible for larger scales of data as applicable to large law firms. Although studies like \citet{de2025comparing}, \citet{wang2022d2gclf}, \citet{watson2023using} present interesting findings, all suffer from the requirement of metadata with the corpus of documents. In real-world scenarios, even major law firms do not possess verified, well-labelled metadata, which narrows the scope of applicability.

\begin{figure}[h]
    \centering
    \includegraphics[width=\linewidth]{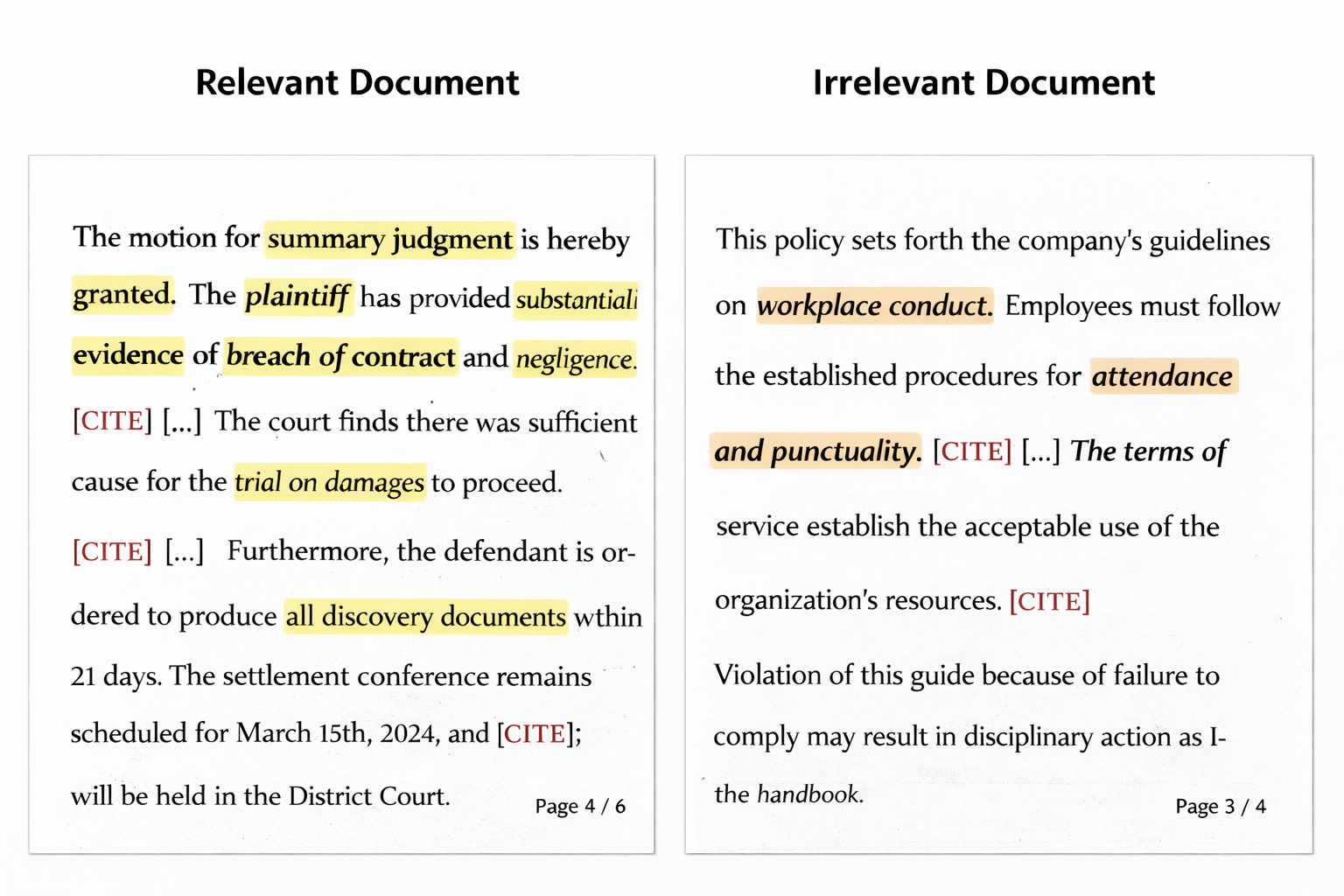}
    \caption{An example of the keywords found in excerpts of relevant and irrelevant documents. The `[CITE]' placeholder is left behind during clause filtering. The relevant document has stronger signals of relevance appearing more often than the irrelevant document.}
    \label{fig:ke-ex}
\end{figure}

To counter these limitations, the authors model classification as a discriminative task driven by phrase-level signals. This is centred on the idea that relevance in legal corpora can be determined by the presence (or absence) of highly indicative phrases \cite{ashley2017artificial} (Fig. \ref{fig:ke-ex}), rather than a broad similarity across the document. By explicitly modelling contrastive phrase-level signals, ReLeVAnT captures lexical markers that reliably and consistently distinguish court filings from non-filing documents, while avoiding reliance on metadata, document structure, or expensive representation learning, often facilitated by LLMs.

ReLeVAnT is a discriminative phrase-driven classification framework for legal documents that leverages contrastive scoring to extract phrases strongly associated with relevant filings. These signals are combined with normalised lexical features and entity filtering, then classified by a shallow neural network, yielding a cost-efficient, structure-independent alternative to computationally intensive, time-intensive embedding and LLM-based methods.
The contributions of this work can be summarised as:
\begin{itemize}
    \item Introducing a novel, lightweight, structure and metadata independent legal document classifier.
    \item Modelling relevance classification as a contrastive phrase-signal task, based on document-level and corpus-level frequency.
    \item Demonstrating 98.7\% F1 score on the LexGLUE dataset, showcasing considerably better performance at much lesser cost.
\end{itemize}

The rest of the paper is structured as follows: Section \ref{RW} discusses related work pertaining to the given problem statement and their contributions and drawbacks. Section \ref{Method} describes the proposed methodology in detail, including the pipeline's intricate design choices. Section \ref{Exp} details the experimental setup and choice of experiments to validate the architecture. Section \ref{Results} demonstrates the impressive results and justifies the choices of hyperparameters and configuration. Section \ref{conc} discusses the potential extension of this work and further avenues to be taken to advance the domain of legal research.

\section{Related Works}
\label{RW}


\subsection{Classical Text Classification \& Retrieval Methods}

Heuristics central to ReLeVAnT are grounded in proven foundations in the Language Modelling space. Term-Frequency signals for document analysis were popularised by \cite{sparck1972statistical}. The key idea of combining term frequency and term rarity is central to ReLeVAnT as well; however, TF-IDF weighs terms globally, whereas the proposed method explicitly contrasts classes. Moreover, classical TF-IDF weighting increases with term frequency but does not explicitly model diminishing returns for repeated occurrences of the same term, which is especially important in longer documents, especially in the legal domain. 

Best Matching 25 (BM25, \citet{robertson2009probabilistic}) is an effective, keyword-based ranking algorithm used in search engines and information retrieval systems to rank documents by relevance to a query. As a descendant of TF-IDF, it improves ranking by calculating term frequency (TF) and inverse document frequency (IDF) while normalising for document length. Like ReLeVAnT, BM25 analyses relevance based on lexical matching and leverages term-frequency importance for scoring. Here, relevance is determined by a query, whereas ReLeVAnT implicitly models class-dependent relevance and does not require external queries.

The work of \citet{cortes1995support} in developing Support Vector Machines (SVMs) has allowed for consistent and reliable classification of high-dimensional features. Like SVMs, ReLeVAnT assumes linear separability in data, but instead of learning weights implicitly, it pre-selects discriminative features, referred to as keywords, explained in Section \ref{Method}. This feature selection prior to classification reduces noise and allows for interpretability.

In recent literature, contrastive learning is generally used to learn common, semantically meaningful features across modalities \cite{radford2021learning} or to learn via the distinction in data and noise, as leveraged in Noise Contrastive Estimation (NCE), \citet{matsuda2021information}). Although NCE assumes similar learning schemes to differentiate between meaningful data and noise, it trains Energy-Based Models (EBMs) probabilistically. In contrast, ReLeVAnT uses contrastive scoring as a lightweight heuristic to isolate discriminative phrases.

To summarise, the central assumptions and concepts that underpin ReLeVAnT are inspired by established work in the Language Modelling space. However, at the time of writing this paper,  no publicly available method implements explicit discriminative filtering to align lexical triggers with legal relevance, while effectively learning features that encode discrimination for classification.


\subsection{Legal Relevance and Document Filtering}

The problem of identifying relevant documents within large legal corpora has been studied in the context of Technology-Assisted Review (TAR) and the TREC Legal Track \cite{cormack2010overview, grossman2011overview}. These works formalise relevance as a function of responsiveness to a given legal matter, typically requiring systems to rank documents based on their probability of relevance.

The TREC Legal Track established standardised benchmarks for evaluating large-scale legal document filtering, demonstrating that classical text-based approaches, including retrieval models and shallow classifiers, can achieve strong performance when calibrated appropriately. However, these frameworks are inherently query-driven, requiring a predefined information need or topic description to guide relevance estimation, introducing dependency.

Subsequent work by \citet{cormack2015autonomy} introduced Continuous Active Learning (CAL), which iteratively refines relevance predictions through human-in-the-loop feedback. While CAL achieves high recall and efficiency in e-discovery settings, it relies on iterative labelling, seed queries, and interactive training workflows, which make it operationally complex.

In contrast, ReLeVAnT formulates legal relevance as a fixed classification problem, independent of external queries or iterative feedback. By modelling relevance using discriminative phrase-level signals extracted directly from document text, ReLeVAnT provides a lightweight, fully automated alternative better suited to large, noisy, unstructured, heterogeneous corpora where metadata, queries, or human supervision may be unavailable.


\subsection{Metadata and Structure-based Methods}

In the legal document classification literature, some of the top-performing methods leverage metadata, structured datasets, and curated features. Although these methods demonstrate impressive results, their reliance on metadata availability and structure makes them rigid and unsuitable for real-world scenarios with noisy firm data (e.g., irregular naming conventions and unstructured data).

Notably, filename signals \cite{li2025document} were used in conjunction with TF-IDF and lightweight models. This allowed for a lightweight, cheap solution, similar to ReLeVAnT, but encoding filenames as strong priors ignores document content and fails on noisy names. This limits generalizability across corpora and to unseen data, while reducing deployability in real-world settings.

Similarly, the approach proposed by \citet{sebastiani2002machine} assumes that clean metadata on authors, keywords, publication venues, and tags are provided. This allows for quicker processing and faster classification, but can be inconsistent when metadata is missing or noisy, as is often the case in legal data \cite{ismaylovna2024problems}. ReLeVAnT relies solely on document text, avoiding external dependencies on metadata that may be incomplete, inconsistent, noisy, or completely unavailable.

Legal judgment classification in the animal protection domain was pioneered by \cite{watson2023using}, which uses domain-specific keywords as signals of relevance, as does the proposed method. However, their method relies on structured judgments and header cues, which are not always available or clean in court filings. Moreover, the narrow domain does not allow for generalisation across the legal corpora. Here, ReLeVAnT demonstrates generalizability beyond domain-specific corpora, is independent of structured headers, and does not rely on the implicit formatting of judgments.

While methods like \cite{wang2022d2gclf} model relationships between entities using graphs, they require accurate NER and more document structure than is available in real-world data. The sensitivity to extraction errors and to graph construction highlights the simplicity of ReLeVAnT and its NER filtering, which is merely a lightweight entity removal mechanism.

\subsection{Deep Learning-based methods}

The advent of deep learning, especially autoregressive methods, has found application in legal text classification. However, a common problem with these methods is the computational and data costs, factors that ReLeVAnT is largely independent of. Recent works like \cite{limsopatham2021effectively} leverage transformers to model contextual relationships and long-range dependencies, both of which are first, not directly necessary for determining relevance, and incur high computational load and scalability issues.

The classification of SCOTUS documents by \citet{undavia2018comparative} uses Word2Vec embeddings with autoregressive architectures and effectively captures patterns across tokens, like the n-gram extraction in ReLeVAnT. Learning these dense semantic embeddings requires a large, labelled dataset and optimising the global representation. This poses restrictions on topic-based modelling, as opposed to relevance-based, and is only explored for structured opinions by the court. Instead of extracting global semantic representations, ReLeVAnT directly targets the most discriminative signals across the corpora and models lexical signals which influence legal relevance.

The work by \citet{joulin2017bag} uses n-grams with a shallow model for text classification. Although the method has proven very effective, it uses character-level n-grams, which carry much less significance than word-level n-grams, especially from the perspective of legal relevance (complaint \& compliance, dismiss \& dismissal, dismiss \& miss). Moreover, the idea of using discriminatory n-grams is novel to our method.



\section{Methodology}
\label{Method}

\begin{figure*}[h]
    \centering
    \includegraphics[width=\textwidth]{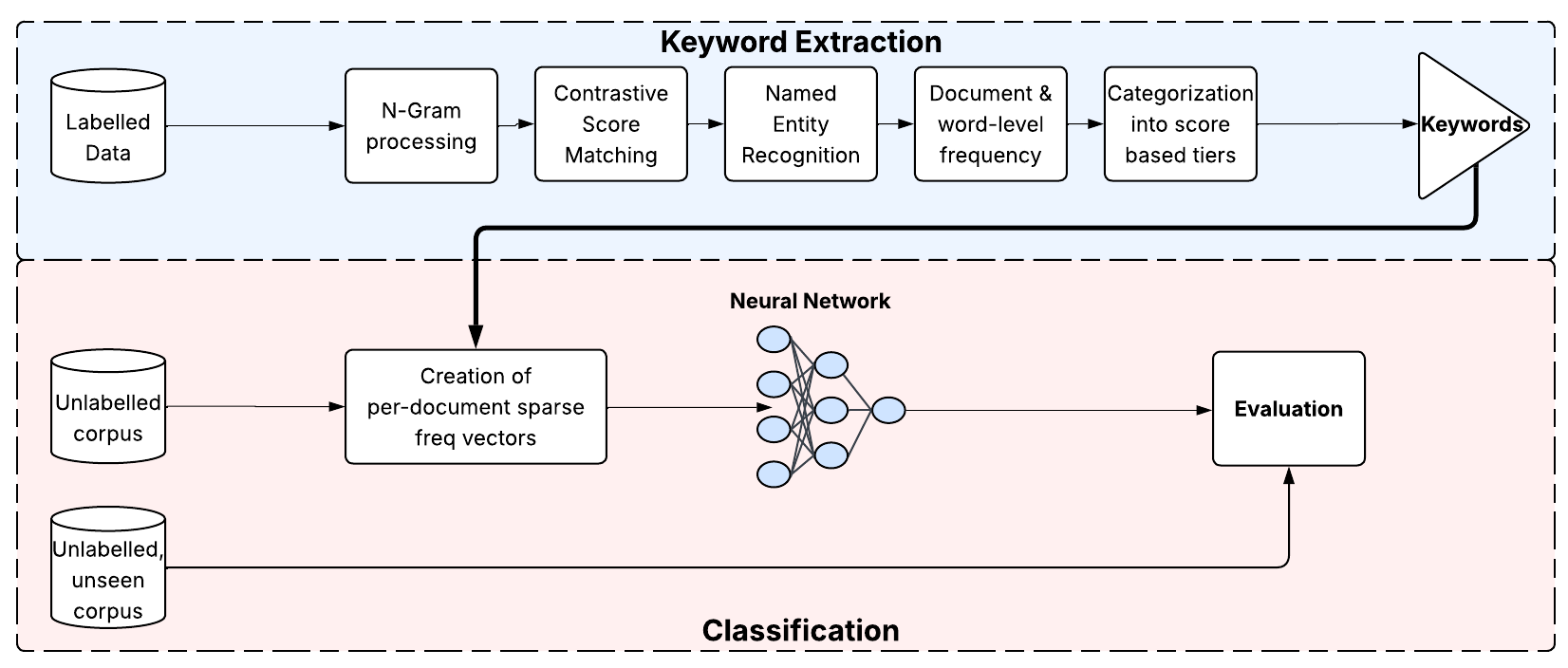}
    \caption{Illustration of the proposed method. The section in Blue highlights the KE stage, and the section in Pink highlights the CLS stage. The given neural network is only for visual purposes, and its exact architecture is detailed in Section \ref{Exp}.}
    \label{fig:model-diag}
\end{figure*}

This section details the proposed methodology of ReLeVAnT. Building on the idea that discriminatory words are strong signals of relevance, the authors propose a 2-stage pipeline comprising the Keyword Extraction pipeline and the Neural Network-based classification pipeline (CLS).

As shown in Fig. \ref{fig:model-diag}, the Keyword Extraction pipeline (KE) begins with NER-based filtering \cite{nadeau2007survey}. This removes all person names, but preserves tags of locations, clauses and organisations. This is particularly important in longer documents where entity names, such as names of plaintiffs, defendants, petitioners, respondents, appellants, appellees, occur very often, and can affect the keywords due to the extraction being frequency-based, as explained further. The remaining clauses and constitutional references are removed to filter out any residual entities left by NER. Then, n-grams are constructed from these filtered texts. These n-grams are key for representing relevance since the signal "Motion to Dismiss" is stronger than individual words like "Motion" or "Dismiss". With these extracted n-grams, contrastive scores are computed. This can be formalised as:
\begin{equation}
    \text{CSM}(t) = \frac{f_t^{+}}{f_t^{+} + (f_t^{-})^{p} + \epsilon}
    \label{eq:csm}
\end{equation}
where \(t\) is the candidate term (extracted n-gram), \(f_t^+\) is the total frequency of \(t\) across all relevant documents, \(f_t^-\) is the total frequency of \(t\) across all irrelevant documents, \(\epsilon\) is the smoothing constant and \(p\) is the penalty exponent applied to negative frequency to control the aggressiveness of punishing terms also in irrelevant docs. Finally, \(\text{CSM}(t)\) is the contrastive score for \(t\) in the range [0, 1], where 1 is highly specific to relevant documents, and 0 is common in irrelevant documents. Using these scores, the document-level frequency is computed. This is formalised as:
\begin{equation}
    \text{DF}(t) = 
        \dfrac{r_t^{+}}{r_t^{+} + r_t^{-} + \epsilon}
\end{equation}
where \(r_t^{+}\) = \(d_t^{+}\) / {\(N^{+}\) represents the fraction of relevant documents containing the term \(t\), \(r_t^{-}\) = \(d_t^{-}\) / {\(N^{-}\) is the fraction of irrelevant documents containing the term \(t\), \(d_t^{+}\) and \(d_t^{-}\) are the number of positive and negative documents containing \(t\), and \(N^{+}\) and \(N^{-}\) are the total number of positive and negative documents in the corpus. Moreover, a hard filter is implemented to reject terms that appear in any irrelevant document. DF(t) counts the fraction of relevant vs irrelevant documents containing the term \(t\). This captures terms that might appear many times in a few irrelevant documents (inflating the negative frequency), but are actually spread across many relevant documents. Here, it should be noted that CSM(\(t\)) does not indicate absence, but rather a lack of lexical signal related to legal relevance in that case.

The combined score is computed by averaging the term-level CSM scores and document-level DF scores.
\begin{equation}
    \text{S}(t) = 
        w \cdot CSM(t) + (1-w) \cdot DF(t)     
\end{equation}
where \(w\) is  the weighing hyperparameter discussed in the next section. This combined how `often' and how `widely' a term \(t\) appears in the corpus. This holistic score determines the absolute importance of \(t\) across the entire corpus.

This combined score is weighted by the frequency of the specific keyword to determine the value of the respective index in the resulting vector. This is expressed as:

\begin{equation}
    \Vec{x_{j}} = S((t_{j}) \cdot count(t_{j}, doc)
\end{equation}
where $\Vec{x}_{j} = \operatorname{count}(t_{j}, \mathrm{doc})$.

These scores are used to construct a (sparse) vector per document, which is then used by a neural network to classify documents as relevant or irrelevant. For a given document \(d\) and a set of \(K\) keywords \({t_1, t_2, ..., t_K}\), the feature vector is formally defined as:
\begin{equation}
    \mathbf{v}(d) = \begin{bmatrix} x_1(d) \\ x_2(d) \\ \vdots \\ x_K(d) \end{bmatrix}
\end{equation}
where each dimension is the raw frequency count $\quad x_j(d) = \text{count}(t_j, d)$.

These feature vectors are constructed for each document, and the documents are classified by a simple feed-forward neural network with a sigmoid activation.

The construction of these feature vectors using a well-weighted, relevance-focused discriminative frequency paradigm ensures accurate and reliable classification.
\section{Experiments}
\label{Exp}

To validate the proposed method, the LexGLUE \cite{chalkidis2022lexglue} benchmark was used. From the tasks in the dataset, ECtHR A, SCOTUS, EUR-LEX, and UNFAIR-ToS were used, with ECtHR A and SCOTUS labelled as relevant due to their direct relevance to European court cases and US Supreme Court opinions, respectively. EUR-LEX and UNFAIR-ToS were labelled as irrelevant due to their nature pertaining to European legislation and Terms of Service clauses, respectively. The CaseHOLD and LEDGAR subsets were excluded from this experiment because they are QA-task-centric and contract-provision-centric, respectively. This dataset was split into train, val, and test sets, consisting of 29532 documents (201666 pages), 9675 documents (74538 pages), and 9007 documents (85441 pages), respectively. Of these, the train set has a relevant-irrelevant ratio of 47\%-53\%, the validation set has a ratio of 25\%-75\%, and the test set has a ratio of 27\%-73\%. These ratios represent realistic scenarios where law firms often have a lot more irrelevant data due to discovery documents, bills and news reports, which cause an imbalance. However, our method is invariant to class imbalance and achieves excellent performance in both situations, as shown in Table \ref{tab:res-class-imba}. To evaluate performance, accuracy and F1 score is utilised. All experiments are conducted using the LexGLUE dataset on HuggingFace and run on an Intel Core Ultra 9 H-class CPU. As explained in Section \ref{Method}, eyecite \cite{eyecite} is used to remove clauses and references. Moreover, the SpaCy core-small model is used for entity filtering. A smoothing constant of 0.01 is used in CSM.

\begin{table}[]
    \centering
    \begin{tabular}{c|c|c}
        \textbf{Class Split} & \textbf{Accuracy}$\uparrow$ & \textbf{F1}$\uparrow$ \\ \hline
        Realistic & 99.3 & 98.7 \\
        Original & 98.9 & 98.1\\  
    \end{tabular}
    \caption{ReLeVAnT's invariance to class imbalance seen in consistent results between less and more aggressive class imbalance scenarios. The `Realistic' scenario has a relevant-irrelevant class ratio of 47\%-53\%, whereas the `Original' ratio in the dataset is 19\%-81\%.}
    \label{tab:res-class-imba}
\end{table}

ReLeVAnT achieves state-of-the-art (SOTA) performance on the LexGLUE dataset, surpassing methods such as majority class selection, always-positive selection, and manually selected keywords, as shown in Table \ref{tab:res-comps}.

\begin{table}[]
    \centering
    \begin{tabular}{c|c|c}
        \textbf{Method} & \textbf{Accuracy}$\uparrow$ & \textbf{F1}$\uparrow$ \\ \hline
        Majority selection & 73.4 & 0 \\
        Always positive selection & 26.6 & 42.1 \\
        Manually chosen keywords & 83.7 & 73.7 \\
        ReLeVAnT & 99.3 & 98.7
    \end{tabular}
    \caption{Comparison of results between ReLeVAnT with and without lemmatisation and Stemming. The \(-LS\) suffix indicates ReLeVAnT with lemmatisation and stemming at KE and CLS.}
    \label{tab:res-comps}
\end{table}

Each intricacy of ReLeVAnT is supported by thorough experiments and solid intuition. The decision to exclude lemmatisation and stemming as preprocessing methods is explained in Table \ref{tab:res-L&S}.

\begin{table}[]
    \centering
    \begin{tabular}{c|c|c}
        \textbf{Method} & \textbf{Accuracy}$\uparrow$ & \textbf{F1}$\uparrow$ \\ \hline
        ReLeVAnT-LS & 95.8 & 92.1 \\
        ReLeVAnT & 98.1 & 96.4 \\
    \end{tabular}
    \caption{Comparison of results between ReLeVAnT with and without lemmatisation and Stemming. The \(-LS\) suffix indicates ReLeVAnT with lemmatisation and stemming at KE and CLS.}
    \label{tab:res-L&S}
\end{table}

Similarly, the n-gram range is thoroughly investigated in Table \ref{tab:res-ngram}, where the best configuration of 4-grams is decided.

\begin{table}[]
    \centering
    \begin{tabular}{c|c|c|c}
        \textbf{N-gram} & \textbf{\#keywords} & \textbf{Accuracy}$\uparrow$ & \textbf{F1}$\uparrow$ \\ \hline
        2-gram & 3322 & 97.9 & 96 \\
        3-gram & 2958 & 98 & 96.3 \\
        4-gram & 2965 & 98.1 & 96.4 \\
        5-gram & 2966 & 98.1 & 96.4 \\
        6-gram & 2966 & 98.1 & 96.4 \\
        7-gram & 2966 & 98.1 & 96.3 \\
        15-gram & 2966 & 98.1 & 96.4 \\
    \end{tabular}
    \caption{Comparison of results of ReLeVAnT across several n-gram ranges. This is a key hyperparameter in the KE pipeline.}
    \label{tab:res-ngram}
\end{table}

The choice of Minimum Term Frequency (MTF) is made from extensive evaluation, as evident in Table \ref{tab:res-MTF}. Here, the optimal choice is at least 30 keywords.

\begin{table}[]
    \centering
    \begin{tabular}{c|c|c|c}
        \textbf{MTF} & \textbf{\#keywords} & \textbf{Accuracy}$\uparrow$ & \textbf{F1}$\uparrow$ \\ \hline
        10 & 3533 & 98.4 & 97.1 \\
        20 & 3533 & 98.4 & 97 \\
        30 & 3533 & 98.4 & 97 \\
        50 & 2966 & 98.1 & 96.4 \\
        100 & 1458 & 97.4 & 95 \\
        250 & 449 & 97 & 94.2 \\
        400 & 220 & 96.3 & 92.7 \\
        1000 & 40 & 93.4 & 86.5 \\
        5000 & 3 & 87.9 & 70.8 \\
    \end{tabular}
    \caption{Comparison of results of ReLeVAnT across MTF thresholds. This is a key hyperparameter in the KE pipeline.}
    \label{tab:res-MTF}
\end{table}

Similar experiments are conducted for the penalty exponent (explained in \eqref{eq:csm}) in Table \ref{tab:res-PX}. This choice of 10 is central to the scoring and affects the pipeline's final performance.

\begin{table}[]
    \centering
    \begin{tabular}{c|c|c|c}
        \textbf{PenEx} & \textbf{\#keywords} & \textbf{Accuracy}$\uparrow$ & \textbf{F1}$\uparrow$ \\ \hline
        1 & 3533 & 98.4 & 97 \\
        2 & 3533 & 98.4 & 97 \\
        10 & 3533 & 98.6 & 97.3 \\
        50 & 3533 & 98.5 & 97.1 \\
        Infinite & 3533 & 98.1 & 96.4 \\
        \end{tabular}
    \caption{Comparison of results of ReLeVAnT across various penalty exponent thresholds. This is a key hyperparameter in the KE pipeline.}
    \label{tab:res-PX}
\end{table}

The document frequency weight hyperparameter is also investigated to balance the weight between document-level and cross-document presence, as shown in Table \ref{tab:res-docFreq}.

\begin{table}[]
    \centering
    \begin{tabular}{c|c|c|c}
        \textbf{DocFreq} & \textbf{\#keywords} & \textbf{Accuracy}$\uparrow$ & \textbf{F1}$\uparrow$ \\ \hline
        0.25 & 3533 & 98.1 & 96.4 \\
        0.33 & 6228 & 98.9 & 97.9 \\
        0.5 & 9140 & 99.2 & 98.5 \\
        0.66 & 9145 & 99.3 & 98.6 \\
        0.75 & 9146 & 99.3 & 98.7 \\
        0.9 & 9154 & 98.9 & 98 \\
        \end{tabular}
    \caption{Comparison of results of ReLeVAnT across various document frequency weights. This is a key hyperparameter in the KE pipeline.}
    \label{tab:res-docFreq}
\end{table}

Finally, variants of the neural network architecture are selected in Table \ref{tab:res-NN} to determine the optimal configuration.

\begin{table}[]
    \centering
    \begin{tabular}{c|c|c}
        \textbf{NN Architecture} & \textbf{Accuracy}$\uparrow$ & \textbf{F1}$\uparrow$ \\ \hline
        A1 & 99.3 & 98.7 \\
        A2 & 98.6 & 97.5 \\
        A3 & 26.6 & 42.1 \\
        \end{tabular}
    \caption{Comparison of results of ReLeVAnT across various NN models. This is a key factor in the CLS pipeline. The architectures in \#nodes for each of A1, A2, and A3 are [512-256-128-64-1], [1024-512-256-128-64-32-1], and [2048-256-64-1].}
    \label{tab:res-NN}
\end{table}

All networks used ReLU for the hidden layers and sigmoid for the output layer (threshold 0.4), with a maximum of 500 iterations, an adaptive learning rate, and early stopping.

To cross-check the validity of the remarkable results, additional experiments were conducted to isolate the highest importance keywords (based on frequency). To assess performance with only the top 3 keywords (`remanded', `testimony', `congressional') and without exactly those 3, and to cross-check whether the neural network is only learning the underlying bias, an experiment with random feature vectors was conducted. These results are evident in Table \ref{tab:res-VAL}.

\begin{table}[]
    \centering
    \begin{tabular}{c|c|c}
        \textbf{Vector Config} & \textbf{Accuracy}$\uparrow$ & \textbf{F1}$\uparrow$ \\ \hline
        Top 3 keywords & 87.9 & 70.8 \\
        Random vectors & 73.4  & 0 \\
        ReLeVAnT-K & 98.4 & 97 \\
        ReLeVAnT & 99.3 & 98.7 \\
        \end{tabular}
    \caption{Comparison of results of ReLeVAnT across specific vector configurations. These experiments are crucial to cross-validate our findings. ReLeVAnT-K refers to the entire keyword list except the top 3 keywords.}
    \label{tab:res-VAL}
\end{table}

\section{Results}
\label{Results}

Experimental results reveal the impressive performance of ReLeVAnT relative to baselines such as majority-class prediction, always-positive selection, and manual keyword selection, as shown in Table \ref{tab:res-comps}.

ReLeVAnT also demonstrates robustness against the relevant class in cases of class imbalance, as shown in Table \ref{tab:res-class-imba}. Here, a difference of 0.4\% in accuracy and 0.6\% in F1 score was observed. This is because the discriminative nature of the keywords selected ensures robustness against imbalances due to near independence of negative frequency at CSM, as explained in Eq. \ref{eq:csm}.

To explore the scope of the vocabulary, the first set of experiments dealt with lemmatisation and stemming. As evident in Table \ref{tab:res-L&S}, ReLeVAnT without this processing step performs considerably better, by upto 2.6\% in accuracy and 4.3\% in F1 score. This can be attributed to the narrowing of vocabulary, leading to a loss of semantic meaning in a term-sensitive domain like Law. For example, `pleading' and 
`plead' is processed as the same word `plead'; however, `pleading' may indicate the current state of a case, whereas `plead' could refer to a clause, a reference, or an event in a news report. Both cases differ in relevance due to the precision of the language used, which is affected by lemmatisation and stemming.

Table \ref{tab:res-ngram} justifies the choice of 4-gram as the optimal choice. As seen here, the performance of ReLeVAnT improves upto the 4-grams, then stagnates near the same value, while the \#keywords also remain the same. This choice allows for optimal performance and scalable results. An experiment with 15-grams is also conducted to test for edge cases of very long repeated sentences, which evidently does not affect performance significantly.

Table \ref{tab:res-MTF} demonstrates optimal performance at MTF 30. Similarly, the performance peaks at 30, with a remarkable accuracy of 98.4\% and an F1 score of 97\%, then diminishes as the MTF threshold rises. This is believed to be due to crucial keywords being missed that may be more `rare' and convey important relevance signals, but may not pass the strict threshold. Hence, an MTF of 30 is chosen.

Table \ref{tab:res-PX} points to the choice of the penalty exponent in the CSM modelling, as explained in Eq. \ref{eq:csm}. Here, a higher penalty exponent points to stricter `punishment' for keywords that occur in negative documents as well. Tuning this to 10 produces the best results, as lower values assign higher scores to keywords that are more prevalent in irrelevant documents, while higher values are `too strict'. With this configuration, 98.6\% accuracy and 97.3\% F1 score are obtained.

Table \ref{tab:res-docFreq} showcases the best performance of ReLeVAnT at 0.75 weight. This indicates that preference for in-document frequency results in better performance over frequency across documents. This reveals an interesting finding: local frequency is more influential than global frequency across the corpus.

Table \ref{tab:res-NN} details the architectures explored and their corresponding results. The structure of 512-256-128-64-1 performs considerably better, showcasing 99.3\% accuracy and 98.7\% F1. A2 and A3 are deeper and shallower networks, respectively, which can lead to overfitting or the curse of dimensionality, significantly reducing performance.

To investigate dependency on top keywords, the authors isolated the top 3 most influential keywords. The exact same setup with just 3 demonstrates passable performance of 87.9\% accuracy and 70.8\% F1 score. This supports our initial assumption of discriminatory words (in this case, `remanded', `testimony', and `congressional') being a strong, separable signal of legal relevance. `Remanded' is present in higher court opinions, a `testimony' is evidence given by a witness under oath, and `congressional' is related to the Congress of the US. All three are intuitively strong markers that would be present only in documents relevant to law. Moreover, removing these 3 to assess the generalizability of the proposed model reveals that the neural network can compensate for the lack of very strong signals and performs very similarly to when the entire keyword list is presented, with only a 0.9\% drop in accuracy and a 1.7\% drop in F1 score. To assess the neural network's randomness, random feature vectors were generated for each document, yielding an exact class split accuracy of 0 and an F1 score of 0. These results further reinforce our hypothesis about the legal relevance and its degree, as measured by the frequency of discriminatory signals.


\section{Conclusion and Future Works}
\label{conc}

This work introduced ReLeVAnT, a lightweight, contrastive, phrase-driven framework that leverages a shallow classifier to achieve state-of-the-art results on binary classification tasks for court filings: relevant vs irrelevant documents. It is completely independent of metadata, filenames, document and directory structure, allowing for massive coverage and robustness at an exponentially lower cost than other methods in the legal text field.

Future work includes multiclass and multilabel classification, as well as downstream tasks. This would enable broader use while tuning the granularity of tasks to find applications across more verticals, possibly outside of law.

\section{Acknowledgment}

We extend our gratitude to \href{https://perssonify.com}{Perssonify} for allocating resources that made this work possible. Moreover, we thank \href{https://zlk.com/}{Levi \& Korsinsky, LLP} for their continued support of our work, and particularly Joseph E. Levi for his valuable inputs and guidance.

\bibliography{custom}

@article{grossman2010technology,
  title={Technology-assisted review in e-discovery can be more effective and more efficient than exhaustive manual review},
  author={Grossman, Maura R and Cormack, Gordon V},
  journal={Rich. JL \& Tech.},
  volume={17},
  pages={1},
  year={2010},
  publisher={HeinOnline}
}

@inproceedings{bhattacharya2021incorporating,
  title={Incorporating domain knowledge for extractive summarization of legal case documents},
  author={Bhattacharya, Paheli and Poddar, Soham and Rudra, Koustav and Ghosh, Kripabandhu and Ghosh, Saptarshi},
  booktitle={Proceedings of the eighteenth international conference on artificial intelligence and law},
  pages={22--31},
  year={2021}
}

@inproceedings{saravanan2008automatic,
  title={Automatic identification of rhetorical roles using conditional random fields for legal document summarization},
  author={Saravanan, Murali and Ravindran, Balaraman and Raman, S},
  booktitle={Proceedings of the Third International Joint Conference on Natural Language Processing: Volume-I},
  year={2008}
}

@inproceedings{chalkidis2022lexglue,
  title={LexGLUE: A benchmark dataset for legal language understanding in English},
  author={Chalkidis, Ilias and Jana, Abhik and Hartung, Dirk and Bommarito, Michael and Androutsopoulos, Ion and Katz, Daniel and Aletras, Nikolaos},
  booktitle={Proceedings of the 60th Annual Meeting of the Association for Computational Linguistics (Volume 1: Long Papers)},
  pages={4310--4330},
  year={2022}
}

@article{pipitone2024legalbench,
  title={Legalbench-rag: A benchmark for retrieval-augmented generation in the legal domain},
  author={Pipitone, Nicholas and Alami, Ghita Houir},
  journal={arXiv preprint arXiv:2408.10343},
  year={2024}
}

@book{ashley2017artificial,
  title={Artificial intelligence and legal analytics: new tools for law practice in the digital age},
  author={Ashley, Kevin D},
  year={2017},
  publisher={Cambridge University Press}
}

@article{katz2017general,
  title={A general approach for predicting the behavior of the Supreme Court of the United States},
  author={Katz, Daniel Martin and Bommarito II, Michael J and Blackman, Josh},
  journal={PloS one},
  volume={12},
  number={4},
  pages={e0174698},
  year={2017},
  publisher={Public Library of Science}
}

@inproceedings{chalkidis2019large,
  title={Large-scale multi-label text classification on EU legislation},
  author={Chalkidis, Ilias and Fergadiotis, Emmanouil and Malakasiotis, Prodromos and Androutsopoulos, Ion},
  booktitle={Proceedings of the 57th annual meeting of the association for computational linguistics},
  pages={6314--6322},
  year={2019}
}

@inproceedings{zhong2020does,
  title={How does NLP benefit legal system: A summary of legal artificial intelligence},
  author={Zhong, Haoxi and Xiao, Chaojun and Tu, Cunchao and Zhang, Tianyang and Liu, Zhiyuan and Sun, Maosong},
  booktitle={Proceedings of the 58th annual meeting of the association for computational linguistics},
  pages={5218--5230},
  year={2020}
}

@inproceedings{de2025comparing,
  title={Comparing Machine Learning and an Expert System for Legal Document Classification},
  author={de Queiroz Santos Filho, Jos{\'e} Jorge and Dantas, Filipe Ara{\'u}jo and da Silva Lima, Melquezedeque and dos Santos, Shirley Barbosa and Genesis, Galileu and da Salva, Maria Gabriely Lima and Pinheiro, {\'A}lvaro Farias and da Silva, Eraylson Galdino},
  booktitle={Conference on Digital Government Research},
  volume={26},
  year={2025}
}

@inproceedings{wang2022d2gclf,
  title={D2GCLF: Document-to-graph classifier for legal document classification},
  author={Wang, Qiqi and Zhao, Kaiqi and Amor, Robert and Liu, Benjamin and Wang, Ruofan},
  booktitle={Findings of the Association for Computational Linguistics: NAACL 2022},
  pages={2208--2221},
  year={2022}
}

@article{watson2023using,
  title={Using machine learning to create a repository of judgments concerning a new practice area: a case study in animal protection law},
  author={Watson, Joe and Aglionby, Guy and March, Samuel},
  journal={Artificial Intelligence and Law},
  volume={31},
  number={2},
  pages={293--324},
  year={2023},
  publisher={Springer}
}

@inproceedings{undavia2018comparative,
  title={A comparative study of classifying legal documents with neural networks},
  author={Undavia, Samir and Meyers, Adam and Ortega, John E},
  booktitle={2018 Federated conference on computer science and information systems (FedCSIS)},
  pages={515--522},
  year={2018},
  organization={IEEE}
}

@inproceedings{li2025document,
  title={Document classification using file names},
  author={Li, Zhijian and Larson, Stefan and Leach, Kevin},
  booktitle={Proceedings of the 2025 ACM Symposium on Document Engineering},
  pages={1--10},
  year={2025}
}

@inproceedings{limsopatham2021effectively,
  title={Effectively leveraging BERT for legal document classification},
  author={Limsopatham, Nut},
  booktitle={Proceedings of the natural legal language processing workshop 2021},
  pages={210--216},
  year={2021}
}

@article{sparck1972statistical,
  title={A statistical interpretation of term specificity and its application in retrieval},
  author={Sparck Jones, Karen},
  journal={Journal of documentation},
  volume={28},
  number={1},
  pages={11--21},
  year={1972},
  publisher={MCB UP Ltd}
}

@article{cortes1995support,
  title={Support-vector networks},
  author={Cortes, Corinna and Vapnik, Vladimir},
  journal={Machine learning},
  volume={20},
  number={3},
  pages={273--297},
  year={1995},
  publisher={Springer}
}

@book{robertson2009probabilistic,
  title={The probabilistic relevance framework: BM25 and beyond},
  author={Robertson, Stephen and Zaragoza, Hugo},
  volume={4},
  year={2009},
  publisher={Now Publishers Inc}
}

@article{matsuda2021information,
  title={Information criteria for non-normalized models},
  author={Matsuda, Takeru and Uehara, Masatoshi and Hyvarinen, Aapo},
  journal={Journal of Machine Learning Research},
  volume={22},
  number={158},
  pages={1--33},
  year={2021}
}

@inproceedings{radford2021learning,
  title={Learning transferable visual models from natural language supervision},
  author={Radford, Alec and Kim, Jong Wook and Hallacy, Chris and Ramesh, Aditya and Goh, Gabriel and Agarwal, Sandhini and Sastry, Girish and Askell, Amanda and Mishkin, Pamela and Clark, Jack and others},
  booktitle={International conference on machine learning},
  pages={8748--8763},
  year={2021},
  organization={PmLR}
}

@inproceedings{cormack2010overview,
  title={Overview of the TREC 2010 Legal Track},
  author={Cormack, Gordon V. and Grossman, Maura R. and Hedin, Bruce and Oard, Douglas W.},
  booktitle={Proceedings of the Text REtrieval Conference (TREC)},
  year={2010},
  organization={National Institute of Standards and Technology (NIST)}
}

@inproceedings{grossman2011overview,
  title={Overview of the TREC 2011 Legal Track},
  author={Grossman, Maura R. and Cormack, Gordon V. and Hedin, Bruce and Oard, Douglas W.},
  booktitle={Proceedings of the Text REtrieval Conference (TREC)},
  year={2011},
  organization={National Institute of Standards and Technology (NIST)}
}

@article{cormack2015autonomy,
  title={Autonomy and reliability of continuous active learning for technology-assisted review},
  author={Cormack, Gordon V. and Grossman, Maura R.},
  journal={arXiv preprint arXiv:1504.06868},
  year={2015}
}

@inproceedings{sebastiani2002machine,
  title={Machine learning in automated text categorization},
  author={Sebastiani, Fabrizio},
  booktitle={ACM Computing Surveys},
  volume={34},
  number={1},
  pages={1--47},
  year={2002},
  publisher={ACM}
}

@inproceedings{joulin2017bag,
  title={Bag of tricks for efficient text classification},
  author={Joulin, Armand and Grave, Edouard and Bojanowski, Piotr and Mikolov, Tom{\'a}{\v{s}}},
  booktitle={Proceedings of the 15th conference of the European chapter of the association for computational linguistics: volume 2, short papers},
  pages={427--431},
  year={2017}
}

@article{nadeau2007survey,
  title={A survey of named entity recognition and classification},
  author={Nadeau, David and Sekine, Satoshi},
  journal={Lingvisticae Investigationes},
  volume={30},
  number={1},
  pages={3--26},
  year={2007},
  publisher={John Benjamins}
}

@article{eyecite,
    title = {eyecite: A Tool for Parsing Legal Citations},
    author = {Cushman, Jack and Dahl, Matthew and Lissner, Michael},
    year = {2021},
    journal = {Journal of Open Source Software},
    volume = {6},
    number = {66},
    pages = {3617},
    url = {https://doi.org/10.21105/joss.03617},
}

@article{ismaylovna2024problems,
  title={Problems of Admissibility and Reliability of Metadata as Evidence. International Journal of},
  author={Ismaylovna, BJ},
  journal={Law and Policy},
  volume={2},
  number={8},
  pages={1},
  year={2024}
}
\end{document}